\documentclass[authorversion, acmsmall]{acmart}

\usepackage{subfigure}  
\usepackage[utf8]{inputenc}  
\usepackage{multirow}  
\usepackage{multicol}  
\usepackage{diagbox}  
\usepackage{bbding}  
\usepackage{xcolor}  
\usepackage{colortbl}  
\definecolor{grey}{RGB}{217,217,214}

\AtBeginDocument{%
  \providecommand\BibTeX{{%
    \normalfont B\kern-0.5em{\scshape i\kern-0.25em b}\kern-0.8em\TeX}}}

\copyrightyear{}
\acmYear{}
\acmDOI{}

\acmBooktitle{- '24: ACM Computing Surveys, April, 2023, XX, XX}
\acmPrice{}
\acmISBN{}

\begin{document}

\title{Federated Computing - Survey on Building Blocks, Extensions and Systems}

\author[R. Schwermer]{Ren\'e Schwermer}
\email{rene.schwermer@tum.de}
\orcid{0000-0002-5750-0538}
\affiliation{%
  \institution{Technical University of Munich}
  \city{Munich}
  \country{Germany}
}

\author[R. Mayer]{Ruben Mayer}
\email{ruben.mayer@uni-bayreuth.de}
\affiliation{%
  \institution{University of Bayreuth}
  \city{Bayreuth}
  \country{Germany}
}

\author[H.-A. Jacobsen]{Hans-Arno Jacobsen}
\email{jacobsen@eecg.toronto.edu}
\orcid{0000-0003-0813-0101}
\affiliation{%
  \institution{University of Toronto}
  \city{Toronto}
  \country{Canada}
}


\begin{abstract}
In response to the increasing volume and sensitivity of data, traditional centralized computing models face challenges, such as data security breaches and regulatory hurdles. Federated Computing (FC) addresses these concerns by enabling collaborative processing without compromising individual data privacy. This is achieved through a decentralized network of devices, each retaining control over its data, while participating in collective computations. The motivation behind FC extends beyond technical considerations to encompass societal implications. As the need for responsible AI and ethical data practices intensifies, FC aligns with the principles of user empowerment and data sovereignty.

FC comprises of Federated Learning (FL) and Federated Analytics (FA). FC systems became more complex over time and they currently lack a clear definition and taxonomy describing its moving pieces. Current surveys capture domain-specific FL use cases, describe individual components in an FC pipeline individually or decoupled from each other, or provide a quantitative overview of the number of published papers. This work surveys more than 150 papers to distill the underlying structure of FC systems with their basic building blocks, extensions, architecture, environment, and motivation. We capture FL and FA systems individually and point out unique difference between those two.

\end{abstract}



\maketitle

\section{Introduction} \label{introduction}
An increase in distributed Internet-of-Things (IoT) devices and the amount of generated data on remote, distributed devices puts stress on central processing entities and the network. Current inductive-based algorithms, such as machine learning training, run primarily on one device (i.e., a GPU), or distributed units execute the computations. In both cases, one entity or stakeholder monitors the entire process. An example is a neural network running on multiple graphical processing units trying to learn patterns in a dataset~\cite{Everitt2016UniversalAI}. However, privacy concerns and regulatory constraints make it more challenging to deploy centralized pipelines due to legal risks, compliance efforts, and a higher consumer sensitivity~\cite{ferrara2018gdprcompliance, basin2018gdprcomplianceII}. Such regularity constraints are put in place all around the world, and they vary in their complexity. Some examples are the General Data Protection Regulation in the European Union~\cite{eu2016gdpr}, the California Consumer Privacy Act in the USA~\cite{california2018ccpa}, and the Personal Data Protection (Amendment) Act in Singapore~\cite{singapore2014pdpa}. One approach to tackle these issues is to shift the computational workload to the devices that generate the data in the first place. With Federated Computing (FC), data scientists and other stakeholders try to disentangle the contradiction between using distributed, privacy-sensitive data, regulatory frameworks, and consumer needs.

However, the literature needs to clearly define FC and how to extend it with approaches from other domains. There are diverse FC paradigms, and it is possible to unintentionally mingle neighboring techniques with FC, such as Federated Databases. FC paradigms are currently Federated Analytics and Federated Learning. However, in the future, we might see additional paradigms emerging as another branch of FC. Those FC paradigms share a similar basis but differ in certain parts. Therefore, it is paramount to have a standard definition for all existing and potential future FC paradigms to distinguish them. Such a framework allows the description of FC systems and subsequently increases comparability with other FC systems to highlight similarities and differences.

Our goal is to show which components (basic building blocks) are at least required to consider something as an FC system and which extensions currently exist. We show which kinds of FC systems exist, which problems they are trying to solve, and which combinations of essential building blocks and extensions the literature currently focuses on. Our proposed framework is expandable to allow it to grow over time or adjust with a changing view of FC systems.

Our contributions are:
\begin{enumerate}
    \item We present a detailed study of recent developments and trends in FC with a focus on the system level. We distinguish in our survey between FL and FA. Our survey highlights research gaps and prevalent system configurations. 
    \item We develop a reference framework for FC and categorize literature accordingly. We highlight how FC basic building blocks and extension work together in different scenarios. A standardized way of describing FC systems allows to compare different works and it enables other researchers to more quickly identify bottlenecks or improvements in the future.
    \item We perform a deep discussion of underlying core technologies of FC on a low-level basis (serilization and communication protocol) and categorize existing FL frameworks. Knowing about the state-of-the art of the underlying communication allows to identify improvements with respect to network traffic and latency.
    \item We present a taxonomy to describe client selection algorithms and summarize widely adopted algorithms.
\end{enumerate}

The rest of this paper is structured as follows. First, we give an overview of other surveys focusing on FC systems in Section~\ref{related_work} and how our paper differs from them. Then, we describe in Section~\ref{federated_computing} how FC works and we discuss its challenges, how to build an FC system, and which external factors influence the design process.

\section{Related Work} \label{related_work}
In this survey, we look at FC from a system perspective. Other surveys followed a similar holistic approach. However, they mainly cover individual components in an FC pipeline separate from each other and focus on Federated Learning (FL) (without covering Federated Analytics (FA)) or a specific domain. We cluster each survey into one of the following categories: Quantitative analysis, tutorial, domain-specific, and taxonomy. Table~\ref{tab:related_work} provides an overview of other surveys and the respective features the capture.

\begin{table}[b]
    \caption{\label{tab:related_work}Summary of existing surveys and what kind of components in an FC system they cover. Those include FC basic building blocks (client selection, aggregation, communication) and extensions from other domains (e.g., privacy-enhancing techniques or compression).}
    \begin{tabular}{l | c c c | c c | c}
    & \multicolumn{3}{c|}{\textbf{FC Basics}} & \multicolumn{2}{c|}{\textbf{FC Extensions}} & \textbf{System Level} \\
    & Client & \multirow{2}{*}{Aggregation} & \multirow{2}{*}{CP} & \multirow{2}{*}{PET} & \multirow{2}{*}{Compression} & \\
    \textbf{Reference} & Selection & & & & & \\    
    \hline
    \rowcolor{grey}
    \multicolumn{7}{l}{Quantitative Survey} \\
    \hline
    Farooq~et~al.~\cite{farooq2021quantitativeSurvey} & & & & \Checkmark & & \\
    Lo~et~al.~\cite{lo2021quantitativeSurvey} &  & \Checkmark &  & \Checkmark &  & \Checkmark \\
    \rowcolor{grey}
    \multicolumn{7}{l}{Tutorial Survey} \\
    \hline
    Aledhari~et~al.~\cite{aledhari2020tutorial} & \Checkmark & \Checkmark & & & & \\
    Yang~et~al.~\cite{yang2019tutorial} & & & & \Checkmark & & \Checkmark \\
    Abreha~et~al.~\cite{abreha2022tutorial} & \Checkmark & \Checkmark & & \Checkmark & \Checkmark & \\
    Li~et~al.~\cite{li2020tutorial} & & \Checkmark & & \Checkmark & & \\
    Kairouz~et~al.~\cite{kairouz2021FLProblems} & & & & \Checkmark & & \Checkmark \\
    Reddy~et~al.~\cite{reddy2023FLTutorial} & \Checkmark & \Checkmark & & & & \\
    Zhang~et~al.~\cite{zhang2021FLTutorial} & & \Checkmark & & \Checkmark & & \\
    \rowcolor{grey}
    \multicolumn{7}{l}{Domain Specific Survey} \\
    \hline
    Xia~et~al.~\cite{xia2021domainFL} & \Checkmark & & & \Checkmark & \Checkmark & \\
    Briggs~et~al.~\cite{briggs2021domainFL} & & \Checkmark & & \Checkmark & \Checkmark & \\
    Zhou~et~al.~\cite{zhou2021domainFL} & & & & \Checkmark & & \\
    Wei~et~al.~\cite{wei2021domainFL} & \Checkmark & & & \Checkmark & & \\
    Thapa~et~al.~\cite{thapa2021domainFL} & & & & \Checkmark & & \\
    Kumar~et~al.~\cite{kumar2021domainFL} & & & & & & \Checkmark \\
    Dirir~et~al.~\cite{dirir2021domainFL} & & & & \Checkmark & & \Checkmark \\
    Enthoven~et~al.~\cite{enthoven2021domainFL} & & & & \Checkmark & & \\
    Pfeiffer~et~al.~\cite{pfeiffer2023domainFL} & \Checkmark & \Checkmark & & & & \\
    Gecer~et~al.~\cite{gecer2024domainFL} & \Checkmark & & & \Checkmark & & \\
    Zhu~et~al.~\cite{zhu2021domainFL} & \Checkmark & \Checkmark & & & & \\
    \rowcolor{grey}
    \multicolumn{7}{l}{Taxonomy Survey} \\
    \hline
    Li~et~al.~\cite{li2021FLSystems} & & & & \Checkmark & \Checkmark & \Checkmark \\
    AbdulRahman~et~al.~\cite{abdulrahman2021FLSystems} & \Checkmark & \Checkmark & & \Checkmark & \Checkmark & \\
    Yin~et~al.~\cite{yin2021FLSurvey} & & & & \Checkmark & & \\ 
    Bellavista~et~al.~\cite{bellavista2022DecentralizedLearning} & & \Checkmark & & \Checkmark & \Checkmark & \\  
    Bonawitz~et~al.~\cite{bonawitz2019FLSystemDesign} & \Checkmark & & & \Checkmark & & \Checkmark \\
    \hline
    Our survey & \Checkmark & \Checkmark & \Checkmark & \Checkmark & \Checkmark & \Checkmark \\
    \multicolumn{7}{l}{} \\
    \multicolumn{7}{l}{CP = Communication Protocol, PET = Privacy-Enhancing Techniques}
    \end{tabular}
\end{table}

A quantitative analysis describes trends and movements in a particular area by evaluating the number of publications. Farooq~et~al.~\cite{farooq2021quantitativeSurvey} and Lo~et~al.~\cite{lo2021quantitativeSurvey} present such an analysis for existing FL research papers without considering FA. They highlight the recent increase in publications starting in 2017. The yearly publications increased from 25 in 2017 to 280 in 2020. Additionally, they cluster papers into different categories to distill focus areas. Most papers investigate the impact of training settings on ML model performance. The main reason to adopt FL is data privacy (62~\% of papers), followed by communication efficiency (23~\%). In general, their research emphasizes the increasing interest in FL. Multiple tutorial-like surveys exist as a reaction, which describe individual components of an FL pipeline together with some application examples.  

Aledhari~et~al.~\cite{aledhari2020tutorial} give in their survey an overview about different FL architectures and their components. Each component with its features and different implementations is highlighted individually. A system overview is missing. Other survey have a similar approach \cite{yang2019tutorial, li2020tutorial, kairouz2021FLProblems, abreha2022tutorial}. Their surveys allow researcher and practitioners new to the field to get a quick overview of existing approaches and algorithms.

Domain-specific surveys investigate the state-of-the art in FL from a use-case perspective~\cite{xia2021domainFL, briggs2021domainFL, zhou2021domainFL}. They focus on Internet-of-Things (IoT) and edge scenarios, among others.

Taxonomy is the practice and science of categorization or classification. A taxonomy is a scheme of classification, especially a hierarchical classification, in which things are organized into groups or types. Li~et~al.~\cite{li2021FLSystems} introduces a taxonomy covering multiple aspects of an FC system. For example, they highlight how to overcome different challenges by illustrating multiple optimization path. They look at each part individually and do not put them into perspective. It is not clear where in the FC system each component is used and how they interact with each other. Additionally, FL specific methods like client selection and aggregation strategies are missing. However, they describe attack vectors on FL training, such as model poisoning and inference attacks. The communication architecture focuses on a high level overview of different options. It neglects the communication protocols used in the back-end of different FL frameworks. A similar taxonomy is introduced by Abdul~Rahman~et~al.~\cite{abdulrahman2021FLSystems}. Their focus is on basic FL building blocks, such as client selection and aggregation algorithms. With respect to communication costs, they briefly mention possible compression techniques to reduce network traffic, but they do not describe it in a broader picture of an FC system. On the application side, they give a comprehensive overview of different use cases and which aggregation algorithms and dataset were used. However, it is not clear which problem each use case was trying to solve (e.g., ML model performance, hardware/network utilization or privacy) and which FL system was deployed. The survey from Yin~et~al.~\cite{yin2021FLSurvey} focus on identifying privacy leakages by introducing a 5W-scenario-based taxonomy. The 5W stands for different questions which give guidance in identifying and resolving attacks. They stand for: "who", "when", "where", and "why". The goal is to identify and quantify security breaches. Bellavista~et~al.~\cite{bellavista2022DecentralizedLearning} introduces a taxonomy which describes decentralized learning systems from a high-level and practitioner point of view with an emphasize on federated environments. It does not provide information about basic FL components such as client selection, aggregation and communication

Another overview of different FL components is given by Bonawitz~et~al.~\cite{bonawitz2019FLSystemDesign}. The focus is on building a centralized FL architecture with fallback aggregators similar to the hierarchical architecture. Besides the given example use case for FL they also describe how FL pipelines work in general and they mention some challenges with suggested solutions.

In our survey, we clearly separate between basic building blocks required to deploy a pure FC system and additional extensions. Besides this new taxonomy, we also add a meta layer, which describes motivation and the hardware environment. After describing each building block and extension separately, we show current trends in FC systems. The reader will have an overview at the end about which FC system is used to achieve which goal, which extensions are mostly used jointly and which FC systems are not yet investigated in sufficient depth. We separate our research and findings into FL and FA systems to highlight differences and shared characteristics.

\section{Federated Computing Definition} \label{federated_computing}
\begin{figure}[t]
    \includegraphics{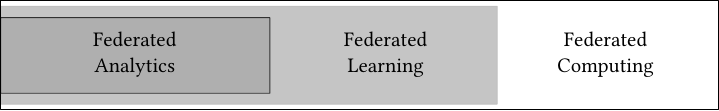}
    \caption{Intersection of definitions for FC, FL, and FA. FC consists of FL and FA, whereas FL systems can contain FA characteristics. FA is a subset of FL and FL is a subset of FC.}
    \label{fig:fc_fl_fa}
\end{figure}

FC belongs to the domain of privacy-preserving computations. It introduces an information asymmetry between the server and clients creating a situation where no one know everything. Its goal is to extract information from distributed data sources without disclosing any raw data. In the process, private data stays on the device. Instead of sending raw data over the network, a node sends computing tasks to participating clients, which execute them and only send the respective computation results (e.g., machine learning model update such as weights and gradients) to the requesting server. An FC round consists of the following steps:

\begin{enumerate}
    \item Server selects participating clients
    \item Server sends computation task to each participating client
    \item Clients execute the computation task and send their individual update to the server
    \item Server aggregates all clients results
    \item For FL: Server distributes the aggregated result back to each client
\end{enumerate}

FC includes FL and FA (see Figure~\ref{fig:fc_fl_fa}). The difference between those two is the type of executed computing tasks and the number of aggregation rounds. FL focuses on machine learning (ML), which mainly consists of multiple aggregation rounds. The goal is to iteratively reduce a loss function and subsequently increase model performance. Conversely, FA leverages statistical operations, such as averages or sums. Each client executes them once, and the server draws conclusions from the data~\cite{wang2022FA}. Some example use cases for FA include model evaluation or debugging~\cite{kairouz2021FLProblems, google2020FA}. Simplified, FL consists of a combination of multiple FA steps~\cite{elkordy2023FA}. FA emerged after FL. It started with Google using it to evaluate the accuracy of Gboard next-word prediction models by using captured data from users’ typing activities on their phones. This is similar to accuracy evaluation~\cite{elkordy2023FA}. Other FA use cases capture analytics for medical studies and precision healthcare or guiding advertisement strategies~\cite{elkordy2023FA}.

Another way to divide FC is by system focus. An FC system can focus on reasoning or learning, translating to FA and FL. Another term for the former is deductive systems and for the latter, inductive systems~\cite{Everitt2016UniversalAI}. Deductive systems are also called "Good old-fashioned AI" and typically rely on rule-based or logical agents~\cite{frankish2014GOFAI, walmsley2012GOFAI}. Conversely, inductive systems try to learn based on the input data and are less prone to changes in the observed environment. An example for FL and FA are the collaborative optimization of an ML model (FL) and its subsequent distributed inference testing on client-side to obtain accuracy metrics for each client (FA). In this survey, we do not cover federated databases. Such systems also have a client-server architecture, which connects distributed databases to one another. The end user only sees one database, even though it consists of multiple ones. Other surveys describe the unique challenges and advantages of Federated Databases (FD)~\cite{azevedo2020federatedDatabases, sheth1990federatedDatabases}. We exclude FD from FC. FDs focus on improving query execution time and increasing availability and reliability of databases. On the other hand FC focuses on managing computations executed on any arbitrary dataset with a focus on privacy. Data in an FD system is openly available to all parties involved whereas FC limits access to client side data.

\begin{figure}[t]
    \includegraphics{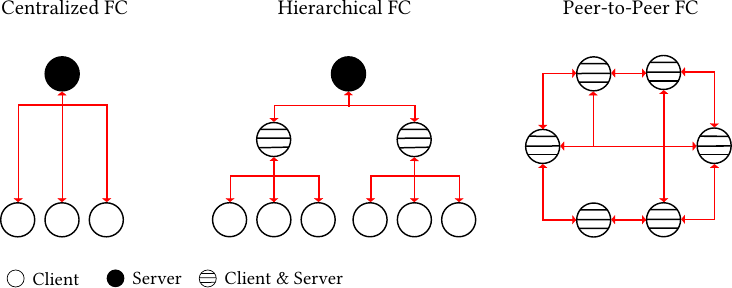}
    \caption{Three FC architectures (centralized, hierarchical and peer-to-peer) with different aggregation server locations. Illustration inspired by~\cite{he2020FedML}. A node can be a client, a server or both, which is indicated by the circle filling.}
    \label{fig:fc_architectures}
\end{figure}

Depending on the FC system architecture, either a central server aggregates all results (central) or clients act as a server as well (peer-to-peer). Figure~\ref{fig:fc_architectures} provides an overview of three FC architectures. The first architecture in the figure shows a centralized FC approach. A server aggregates all results from the clients. The hierarchical architecture has an intermediate layer between clients and servers to increase redundancy. The devices in this intermediate layer act as servers and clients simultaneously. This layer adds robustness to the overall system. It can still generate insights even if one cluster fails. Additionally, it allows to cluster clients by categories, which could also be spatial or user-based. A fully decentralized FC system runs peer-to-peer without a central server. Another term for this architecture is galaxy FC~\cite{hu2020galaxyFL, li2022_123_unknown_manual}. Those systems are complicated to manage and it is challenging to converge to a given performance threshold if the ML optimization per client keep bouncing between multiple states instead of converging to a local or global optimum. All those architectures work for FA and FL.

\subsection{Scenarios} \label{scenarios}
FC systems can have different objectives, run on different devices, or with different data distributions. Table~\ref{tab:fc_scenarios} summarizes commonly used ways to describe those scenarios. We structure them by a guiding question into three groups. All production-ready FC systems consist of a combination of one option per group - for example, a model-centric, horizontal, and cross-device FC system.

\begin{table}[b]
    \caption{\label{tab:fc_scenarios}Three different groups of scenarios organized by a guiding question. Each scenario has a short description. An FC scenario consists of at least one scenario per group, for example model-centric, vertical, cross-silo FC.}
    \begin{tabular}{ l | l | c }
        \textbf{Scenario} & \textbf{Description} & \textbf{Guiding Question} \\
        \hline
        Model-Centric & Curating or improving output & \multirow{2}{*}{Why?} \\
        Data-Centric & Curating or improving input & \\
        \hline
        Horizontal & Same features with different users & \multirow{2}{*}{How?} \\
        Vertical & Different features with same users & \\
        \hline
        Cross-Device & E.g. mobile phones, IoT devices & \multirow{2}{*}{Where?} \\
        Cross-Silo & E.g. hospitals, manufacturing sides &
    \end{tabular}
\end{table}

The first guiding question "Why" addresses the overall objective of the FC scenario. Does the user want to improve an ML model output or generated insights (model-centric), or does the user want to improve a data set and then make it available for others to train on (data-centric)? A model-centric approach tries to extract as much information as possible from a given dataset, e.g., by tuning hyperparameters of the model or tweaking the optimizer. On the other hand, data-centric is a quality-over-quantity approach and focuses on collecting and using only specific data suitable for a particular use case. It is a paradigm emphasizing that systematic design and engineering of data is essential for developing AI-based systems~\cite{jakubik2022dataCentric}. Therefore, performance improvements result from improving the quantity and quality of the data instead of changing the underlying model architecture. FL and FA require data. Gröger emphasizes challenges for data management, governance, and democratization in an industry context, which highlights the potential for data owner to make a data-driven approach more feasible in the future~\cite{groeger2021AIneedsData, Gudivada2017DataQC}. The current focus in the literature and the industry is on model-centric scenarios, as seen in Section~\ref{conclusion}. This approach can start with a pre-trained ML model or from scratch.

The next group of scenarios considers the data distribution of the clients. In horizontal FC (HFC), all participating clients have the same features, but different users. An example is predicting the next words written on Google’s Gboard~\cite{bonawitz2019FLSystemDesign}. The users per client differ, but the features (word predictions) are the same. Vertical FC (VFC) is the opposite. Each client monitors different features, but they share the same user. An example application is the finance sector, where retailers and banks store historical data on the same person but with different features~\cite{yang2019tutorial}.

Lastly, we group each scenario by the location of the participating devices. A cross-device scenario uses distributed devices with a high degree of individual ownership. Those devices could be mobile phones or wearables, such as smartwatches or home assistance systems. Cross-device scenarios work with up to thousands of devices, which all can have a different owner, generally private persons. On the other hand, cross-silo scenarios leverage data allocated on devices or entities with much less diverse ownerships. Some examples come from the healthcare and manufacturing domains. Multiple hospitals or pharmaceutical companies can collaborate and jointly generate results in an FC fashion~\cite{xu2020FLHealthCare, rieke2020FLHealthCare}. The training process still runs on dedicated hardware, which could be mobile phones. Therefore, there needs to be a clear cut between the definition of cross-device versus cross-silo scenarios. Cross-silo is mainly limited to a few hundred participating clients due to the organizational complexity, and the owners of the clients' hardware are businesses.

\subsection{Problems} \label{cha:problems}
FC systems increase data privacy due to limiting data access. However, FC faces a multitude of problems. The assumption for FL is that each client has labelled data and therefore follows a supervised training. We cluster the problems into three categories:
\begin{enumerate}
	\item Improve insights or ML performance,
    \item Improve privacy or security,
    \item Improve hardware or network utilization.
\end{enumerate}

The first problem is mainly due to an uneven distribution of features and labels on the clients. In an FC system, the server has no data access on client side. It only knows some meta information, such as image resolution or for time series the respective units of each column. In an ideal scenario all clients' datasets have similar statistical attributes, which yield similar results of the executed computation tasks as well. However, in real scenarios some clients' datasets might be biased towards certain labels. Therefore, the same execution task can yield different results per client. This is called non-independent and identically distributed (non-IID) data. Aggregating results based on non-IID data is a challenge due to its impact on the final result on server-side. For some cases the aggregation of individual client updates can result in worse results compared to a traditional centralized approach. Choosing a subset of available clients (Section~\ref{client_selection}) or a suitable aggregation strategy (Section~\ref{cha:aggregation_algorithm}) can improve the generated insights. Another issue in this context are clients dropping out during an FC process. This can also result in a non-IID scenario even though clients were properly selected at the beginning of the process. furthermore, the entire run can get delayed, because the server is waiting for all clients to finish their execution task.

The second problem mainly copes with the possibility of attacks on an FC system to infer raw data from the individual clients model and data leakage during the process. In general, there are two types of attacks: Black-box and white-box attacks. In black-box settings, the adversary’s access is limited to the model’s outputs only. The  adversary can query the model with an arbitrary input x and obtains the prediction vector f(x). In white-box settings, the adversary has full access to all components of the model. The access includes the model’s architecture, parameters, and hyper-parameters. Also, the adversary can inspect intermediate computations and prediction vectors. Attacks can occur at different stages (input data, training and inference phase) during an FC round. An attack during the input phase tries to poison the data in such a way that the final model is impaired. This attack originates from a participating client. In the training phase, participating clients can try to infer data based on the updates they get from the server or alter the model on purpose to again impair the final model. Inference attacks happen during the training process or at the end. Their goal is to leak information about the training data and not to impair any data or models.

The third problem is due to the distributed nature of FC systems. Each client sends its updates either to a central server or other clients (peer-to-peer). Reducing this network overhead without interfering with result accuracy is one research area. Additionally, considering each clients individual hardware and data in the scheduling of the training helps to use computational resources more efficiently.

\subsection{Components} \label{building_blocks_and_extensions}
\begin{figure}[b]
    \includegraphics{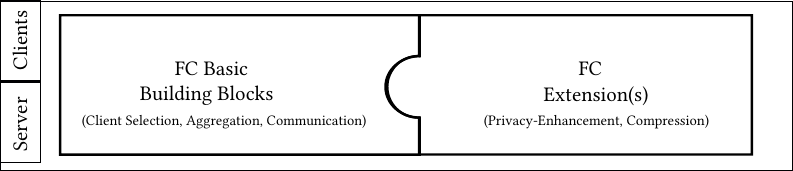} 
    \caption{Interaction between basic building blocks for FC systems and optional extensions. The basic building blocks consist of hardware (devices hosting data set and an aggregation server) and software components (client selection, aggregation strategy and communication protocol).}
    \label{fig:fc_system_overview}
\end{figure}

FC systems consist of different components (basic building blocks), which can be enhanced with extensions from other domains (Figure~\ref{fig:fc_system_overview}). The goal is to a build an FC system to solve one or all above mentioned problems. There are multiple options available for each basic building block and extension. A systems' performance depends on how well all components work together for the respective use case. The following paragraphs shortly describe all required basic building blocks for a FC system and two widely-adopted types of extensions. A detailed description of all components follows in Section~\ref{essential_building_blocks} and Section~\ref{extensions}.

\begin{figure}[t]
    \includegraphics{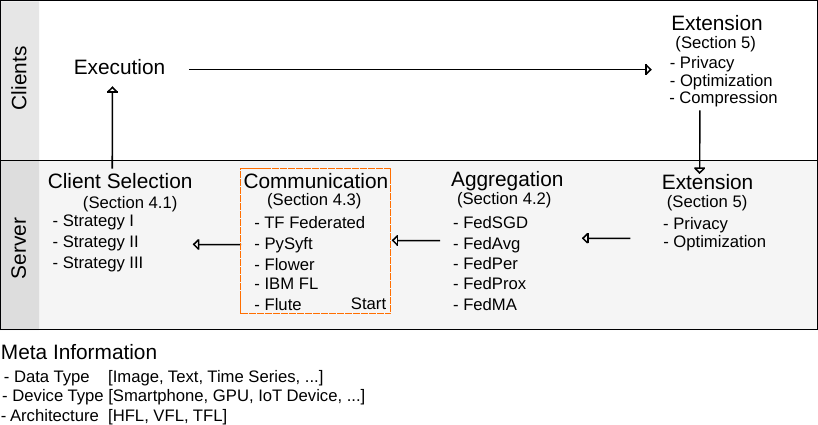}
    \caption{Framework to describe how different building blocks in an FC system work together. The illustrated pipeline focuses on FC basic building blocks and it shows some example options for each block. Additionally, it shows some options for the systems meta information.}
    \label{fig:fc_system_basics}
\end{figure}

We separate those basic building blocks into hardware and software components. The hardware consists of devices hosting private data sets and an aggregation server. For this definition, the devices' computational resources are not of importance. The computational resources can vary from computationally weak edge devices to strong GPU servers. The software part consists of three components: Client selection, aggregation strategy and communication protocol (e.g., gRPC, WebSocket, HTTP). Figure~\ref{fig:fc_system_basics} provides an overview on how these components work together with some options per step. To further improve different aspects of such an FC system it is possible to extend it with methods from other domains. For each building block and extension different options exist.

In the following paragraphs, we shortly explain the three software components individually, before going into detail in their respective section. FC works with distributed clients. The number of participating clients can vary from a few to multiple thousands. Following a quantity over quality approach might interfere with the above three mentioned problem categorize (Section~\ref{cha:problems}). Therefore, it is necessary to introduce constraints on the client selection process. Those constraints can be of different nature and each of them tries to improve at least one problem. For example, taking clients with a similar data set size reduces the idle time of individual clients. This approach decreases the risk of clients dropping out during the training and helps to improve FC performance and hardware utilization. Another example for a client selection constraint is to pick only geographically close clients to reduce network failures. In Section~\ref{client_selection} we go into detail on how different client selection algorithms work and which advantages and disadvantages they have.

Each client sends its results either to a central server or to another client (peer-to-peer). Those results have to be combined to leverage each individual clients' insights. This can either be an average of all results or more complex operations. Some other examples are to compute weighted averages or to cluster clients into different groups. Taking an average is called Federated Averaging (\textit{FedAvg}) and it is one of the first aggregation algorithms~\cite{McMahan2016FL}. The focus of an aggregation algorithm is to improve the generated insights. However, as we see later in Section~\ref{cha:aggregation_algorithm}, some aggregation algorithms are prone to privacy and security risks (e.g., reverse-engineering the raw data with the client updates) or they introduce bottlenecks in the hardware or network utilization.

In an FC system, each node (server or client) has its own unique IP or identifier. During a training round they have to communicate with each other and send data to one another. Engineers leveraged different network protocols and serialization methods in the context of FC. Building an FC system works in theory with all kinds of combinations of communication protocols and serilization methods. In Section~\ref{frameworks_and_network_protocols} we explain briefly how different communication and serilization methods work. Then we introduce currently available FC frameworks with their respective architectures.

\subsection{Meta Layer} \label{cha:meta_layer}
Besides the hardware and software components other more holistic aspects are also important to classify an FC system. We split those aspects into three different categories:
\begin{enumerate}
	\item Problems (see Section~\ref{cha:problems}),
    \item Hardware environment,
    \item Economic feasibility study
\end{enumerate}

In Section~\ref{building_blocks_and_extensions} we already introduced the first meta layer "Motivation" with its three levels of motivation. Improving insights with an FC system mainly faces the potential issue with non-IID data on the clients. In contrast to a centralized approach, it is difficult to balance the distribution of features and labels during an FC run. In the literature the focus is on three different approaches to improve the generated insights. The first one works with avoiding non-IID in the first place by using a suitable client selection process (Section~\ref{client_selection}). Secondly, more and more specialized and tailored aggregation algorithms are developed to balance the clients updates (Section~\ref{cha:aggregation_algorithm}). Lastly, leveraging optimization techniques developed originally for other applications can help to reduce the negative impact of unbalanced data sets on the final results.

The next meta layer considers the environment an FC system runs in. Available options are:

\begin{enumerate}
    \item simulated FC system (clients and server) on one device,
    \item virtual machines~(VM),
    \item physically separated devices or
    \item a combination or hybrid.
\end{enumerate}

Choosing a suitable hardware environment requires a balance between development speed, maintenance and boundary conditions. For example, testing the effect of different client selection algorithms on the final result (e.g., machine learning prediction performance) does not depend on the network bandwidth. Therefore, it is feasiable to simulate all clients and the aggregation server on one device. On the other hand, tuning the way clients communicate with the server or each other requires a realistic emulation of network communication to understand the impact of different approaches on the communication behavior.

Lastly, all FC systems compete with centralized state-of-the-art approaches. Therefore, they have to have advantages to be widely used in real-world applications. Those can be legal, public relations or trust issues. However, all of them boil down to an economical impact for the respective stakeholder. Currently, the focus of the literature is on technical aspects and less on economical feasibility. However, this area is growing and some authors already mapped some existing economical frameworks on FC system to investigate different scenarios. 

\section{Modules}

\subsection{Basic Building Blocks} \label{essential_building_blocks}
The following sections describe each basic building block of an FC system individually in more detail. Those are client selection (Section~\ref{client_selection}), aggregation (Section~\ref{cha:aggregation_algorithm}), and communication (Section~\ref{frameworks_and_network_protocols}). We give an overview of existing approaches and cluster them by different criteria.

\subsubsection{Client Selection} \label{client_selection}
FC systems work with distributed clients and the number of those can range from just a few to multiple thousands of devices. Choosing a suitable pool of clients helps alleviating one or multiple of the challenges introduced in Section~\ref{cha:problems}. Currently, most FC systems work in the order of magnitude of 40 clients. For those cases, the client selection is done manually and it considers either 100~\% of the available clients or randomly picks a subset. However, non-IID data on the clients or geographically separated clients can introduce issues with respect to knowledge gain for example due to  non-converging ML models, slower network traffic or higher latencies. Therefore, we cluster the goal of client selection algorithms into three categorize:

\begin{enumerate}
	\item Decrease training time,
    \item Decrease network traffic,
    \item Improve generated insights.
\end{enumerate}

There are different strategies available to achieve either one or all of the above listed goals. The deployed strategy depends on the number of clients, their computational resources and the available network. Nishio~et~al.~\cite{nishio2019_13_single_resource} proposes \textit{FedCS} and focus on mobile edge devices with limited and heterogeneous computational resources. Clients are selected if their time to run one aggregation round is below a given threshold. This approach does not allow a pre-selection of clients and needs at least one round of training. Abdulrahman~et~al.~\cite{abdulrahman2021_3_multiple_resource} follow a similar approach with \textit{FedMCCS}, which additionally considers CPU, memory and energy constraints. These algorithms are not suitable for FA, which only has one aggregation round. Besides computational resources, researchers select clients depending on their network connection. The experiments of Xu~et~al.~\cite{xu2021_13_unknown_resource} run with clients being connected all to one wireless link. Instead of having a throughput maximization approach they follow an optimization strategy to improve the systems' learning performance under finite bandwidth and energy constraints. It is also possible to have a more fluctuating set of clients which change during FL training. Here, a selection criterion can be the current loss of each client to increase convergence speed of the trained model~\cite{cho2020selectbasedonloss}.

\subsubsection{Aggregation} \label{cha:aggregation_algorithm}

After every round, each participating client sends an update to an aggregation server that merges all updates to a single model. A round refers to either a single event, a batch or an epoch. Such updates can be scalars, vectors, or matrices containing for example an ML model's gradients, weights, or biases for FL systems. An aggregation strategy solely focuses on how to process such updates. The server can apply statistical techniques, such as average or mean, or filter based on thresholds. All those modifications can run on an element-wise order or follow another arbitrary order. Some proposed strategies also change the client selection process. All aggregation strategies aim to address the potential issue with non-IID on the client side. The challenge is generalizing client updates on the server side by simultaneously personalizing the models on the client side. Therefore, improvements for aggregation strategies apply to either the server or client side or both.

\textit{FedSGD} and \textit{FedAvg} are the first aggregation algorithms designed for FL systems~\cite{mcmahan2023communicationefficient}. They run element-wise calculations on the input. The equations for calculating \textit{FedSGD} (Eq.~\ref{eq:fedsgd}) and \textit{FedAvg} (Eq.~\ref{eq:fedavg}) on the server side differ by the number of training rounds on the client. For \textit{FedSGD}, each client takes one step of gradient descent and then it sends its update to the server. The server takes a weighted average of all updates. \textit{FedAvg} differs from \textit{FedSGD} by running more iterations on each client before aggregating the results. The learning rate is given by $\eta$, K is the set of clients, t describes one time step, w is the model, n the number of data points on the client and the Nabla operator $\nabla$ converts a field of scalars to a field of vectors.

\begin{equation}
    \label{eq:fedsgd}
    FedSGD:\quad w_{t+1} \gets w_{t} - \eta\nabla f_{k}(w_{t})\quad \textnormal{with}\quad \nabla f(w_{t}) = \sum_{k=1}^{K} \frac{n_{k}}{n}g_{k}
\end{equation}
    
\begin{equation}
    \label{eq:fedavg}
    FedAvg:\quad w^k \gets w^k - \eta\nabla F_{k}(w^k)
\end{equation}

McMahan~et~al.~\cite{mcmahan2023communicationefficient} introduces three parameters to describe an aggregation strategy. The first is C, representing the fraction of clients participating in the computations on each round. It ranges from 0 to 1, with one referring to all available clients. An FC system consists of at least two clients to enable some form of aggregation and at least some protection against reverse-engineering the raw data based on the model updates. The second parameter is E, which refers to the number of training passes on each client before aggregating their updates. It is an absolute value and refers directly to the number of local training rounds. Third, the parameter B describes the size of the mini-batch in relation to the clients' dataset size. "1" refers to using the entire local dataset as one batch. Besides those three parameters, Arivazhagan~et~al.~\cite{arivazhagan2019fedper} introduced the parameter K. It describes the number of layers of a neural network trained exclusively locally. Those layers do not change after receiving an update from the server.

Tweaking those parameters allows the development of new specialized aggregation strategies designed to work in the dedicated system environment. Table~\ref{tab:aggregation_strategies} overviews various aggregation strategies and their respective parameter settings. Those aggregation strategies solely focus on FL and the challenge with non-IID data on the client side. For example, \textit{FedPer} and \textit{FedDist} focus heavily on neural networks. The former only updates a given number of layers and keeps others local. This approach aims to make the models generalizable for all clients by simultaneously keeping a certain degree of personalization. The latter aggregation strategy calculates the Euclidean distance between neurons in a neural network to identify diverging neurons. Focusing on neural networks eliminates those aggregation strategies from being used in FA systems. 

\begin{table}[b]
    \caption{\label{tab:aggregation_strategies}Classification of FL aggregation strategies depending on C (number of clients participating in the training with one being 100~\%), E (number of training iterations before aggregation), B (local mini-batch size with one referring to the entire local dataset as one batch), and K (number of unchanged / frozen layers).}
    \fontsize{9}{11}\selectfont
    \begin{tabular}{l | c | c | c | c | l}
    & C & E & B & K & \multicolumn{1}{c}{Note} \\
    \hline
    FedSGD~\cite{mcmahan2023communicationefficient} & $\leq$ 1 & 1 & 1 & 0 & - \\
    FedAvg~\cite{mcmahan2023communicationefficient} & $\leq$ 1 & > 1 & $\leq$ 1 & 0 & - \\
    \multirow{2}{*}{FedPer~\cite{arivazhagan2019fedper}} & \multirow{2}{*}{$\leq$ 1} & \multirow{2}{*}{> 1} & \multirow{2}{*}{$\leq$ 1} & \multirow{2}{*}{> 0} & Clients update only a subset of NN layers \\
    & & & & & and train the other layers locally \\
    FedProx~\cite{li2020FedProx} & $\leq$ 1 & Varies per client & $\leq$ 1 & 0 & - \\
    FedMA~\cite{wang2020FedMA} & $\leq$ 1 & > 1 & $\leq$ 1 & 0 & - \\
    FedAT~\cite{chai2021FedAT} & $\leq$ 1 & > 1 & $\leq$ 1 & 0 & Clusters clients based on latency \\
    FedDane~\cite{li2020FedDane} & Varies per round & > 1 & $\leq$ 1 & 0 & - \\
    FedZIP~\cite{malekijoo2021fedzip} & $\leq$ 1 & > 1 & $\leq$ 1 & 0 & Compresses updates \\
    \multirow{2}{*}{FedDist~\cite{rk2021FedDist}} & \multirow{2}{*}{$\leq$ 1} & \multirow{2}{*}{> 1} & \multirow{2}{*}{$\leq$ 1} & \multirow{2}{*}{0} & Calculates euclidean distance to \\    
    & & & & & identify diverging neurons \\
    \multirow{2}{*}{FedMAX~\cite{chen2020fedmax}} & \multirow{2}{*}{$\leq$ 1} & \multirow{2}{*}{> 1} & \multirow{2}{*}{$\leq$ 1} & \multirow{2}{*}{0} & Max entropy regularization to equalize  \\
    & & & & & activation vectors in an NN layer \\
    \end{tabular}
\end{table}

Besides \textit{FedDane}, all other aggregation strategies have a pre-defined set of clients at the beginning of the first FL training round. However, a pre-defined set of clients is still prone to clients dropping out during training. The aggregation server might replace those clients with new ones. Therefore, the set might change over time, but the defined set is not a crucial part of the aggregation strategy. At the same time, \textit{FedDane} incorporates the change of selected clients into its core structure. It approximates the gradients using a subset of gradients from randomly sampled clients~\cite{li2020FedDane}. It achieves theoretically better results than \textit{FedAvg}, but it underperforms in actual experiments and requires double the number of communication rounds due to adjusting the clients based on an optimization problem.

Conversely, \textit{FedProx} varies the number of local training iterations per client before aggregating the results. \textit{FedProx} differs from \textit{FedAvg} by allowing for a variable number of training iterations on the client side based on their available dataset and computational resources. This approach results in some clients training more rounds than others. The server aggregates those partial solutions~\cite{li2020FedProx}. The parameters of the aggregation strategy for \textit{FedAT} are the same as for \textit{FedAvg}. However, it clusters the available clients based on latency to improve the overall training time and test accuracy by having clients with similar time and dataset constraints~\cite{chai2021FedAT}.

\subsubsection{Communication} \label{frameworks_and_network_protocols}
FC systems work with remote clients. It describes a distributed client and server architecture with data being transferred between those nodes. This section describes FL frameworks and their principles in more detail. 

Multiple organisations, institutes or other stakeholders develop FL frameworks. Some of them have a specific focus on a certain domain, some are not for commercial use and others are not further developed. Several frameworks have been developed, such as \textit{PySyft} from openMined~\cite{ryffel2018generic}, \textit{TensorFlow Federated} from Google~\cite{bonawitz2019FLSystemDesign}, \textit{IBM FL} from IBM~\cite{ludwig2020ibm}, \textit{FedAI/FATE} from WeBank~\cite{webank2018flwhite}, \textit{Clara SDK} and \textit{FLARE} from Nvidia~\cite{tetreault2020nvidia, flTeam2021nvidia}, \textit{FedML}~\cite{he2020FedML}, \textit{Paddle FL} from Baidu, \textit{Fed-BioMed}~\cite{inria2021fedmed}, \textit{Flower}~\cite{beutel2021flower}, \textit{FLUTE} from Microsoft~\cite{dimitriadis2022flute}, \textit{Substra} from Owkin~\cite{owkin2019FLFramework}, \textit{OpenFL} from Intel~\cite{foley2022FLFramework}, \textit{FederatedScope} from Alibaba~\cite{xie2022federatedscope} \textit{APPFL} from the Argonne National Laboratory (USA)~\cite{ryu2022appfl}, and \textit{Vatange6}~\cite{smits2022vantageFL}. \textit{LEAF}~\cite{caldas2019leaf} provides tools to benchmark different pre-selected models in a FL setting. Karimireddy~et~al.~\cite{karimireddy2023FLFrameworkSummary} assess the strengths and weaknesses of 14 different FL frameworks, ranging from supported data distributions and communication topologies to available built-in advanced privacy and security features. 

\begin{table}[b]
    \caption{\label{tab:fl_frameworks}FC frameworks with their communication protocols and serialization method(s). All frameworks state in their white papers or documentation a focus on FL. However, they might be adjustable to FA use cases.}
   \fontsize{9}{11}\selectfont
    \begin{tabular}{l | c c c c | c c c c}
           & \multicolumn{4}{c}{\textbf{Protocol}} & \multicolumn{4}{c}{\textbf{Serialization}} \\
           & gRPC & WebSocket & HTTP & GLOO & Pickle & JSON & Protobuf & FOBS \\
    \hline
    APPFL~\cite{ryu2022appfl} & \Checkmark & & \Checkmark & & \Checkmark & & \Checkmark & \\
    FedBioMed~\cite{inria2021fedmed} & \Checkmark & & & & & \Checkmark & \Checkmark & \\
    FedN~\cite{ekmefjord2021fedn} & \Checkmark & & \Checkmark & & & \Checkmark & \Checkmark & \\
    FedScope~\cite{xie2022federatedscope} & \Checkmark & & & & & & \Checkmark & \\
    Flower~\cite{beutel2021flower} & \Checkmark & & & & & & \Checkmark & \\
    Flute~\cite{dimitriadis2022flute} & & & & \Checkmark & \Checkmark & & & \\
    IBM FL~\cite{ludwig2020ibm} & & & \Checkmark & & \Checkmark & \Checkmark & & \\
    FLARE~\cite{tetreault2020nvidia} & \Checkmark & & & & & & \Checkmark & \Checkmark \\
    OpenFL~\cite{foley2022FLFramework} & \Checkmark & & & & & & \Checkmark & \\
    PySyft~\cite{ryffel2018generic} & & \Checkmark & & & & & \Checkmark & \\
    TFF~\cite{bonawitz2019FLSystemDesign} & \Checkmark & & & & & & \Checkmark & \\
    Vantage6~\cite{smits2022vantageFL} & & \Checkmark & \Checkmark & & & \Checkmark & & \\
    \end{tabular}
\end{table}

We divide the required components into communication protocols and serialization methods. Serializations methods are either binary or contextual. The former converts an input into a series of bytes and the latter uses data formats such as JSON or XML to transfer information. Contextual serialization contains details about the data's structure and purpose, making it simple for a human reader to interpret and comprehend. Table~\ref{tab:fl_frameworks} provides an overview of FC frameworks with their respective components. Not all frameworks state exactly which protocol and serialization methods they are using and instead refer to a vague statement in their documentation saying that messages are sent over the internet. All frameworks are consistently promoted for FL. However, they can be adjusted to also deploy an FA system. Eight out of twelve frameworks use \textit{gRPC} in combination with the binary serialization Protobuf. The \textit{gRPC} protocol has no browser support. \textit{WebRTC} and \textit{WebSockets} support browser integration and are therefore used for \textit{PySyft} with its browser based Duet implementation. Other outliers are the FL frameworks of IBM and Microsoft, which use \textit{HTTP} and \textit{GLOO} as their respective network protocol. It is not recommended to use Pickle serialization in environments with untrusted parties due to potential security issues. The documentation of Pickle emphasizes the fact that it is possible to construct malicious pickle data which could execute arbitrary code during unpickling~\cite{pickle2023documentation}. Instead, they recommend to either use \texttt{hmac} for message authentication in Python or to switch to JSON serialization. \textit{IBM FL} offers both serialization methods.

\subsection{Extensions} \label{extensions}
\subsubsection{Privacy Enhancement} \label{privacy_enhancement}

\begin{figure}[b]
    \includegraphics{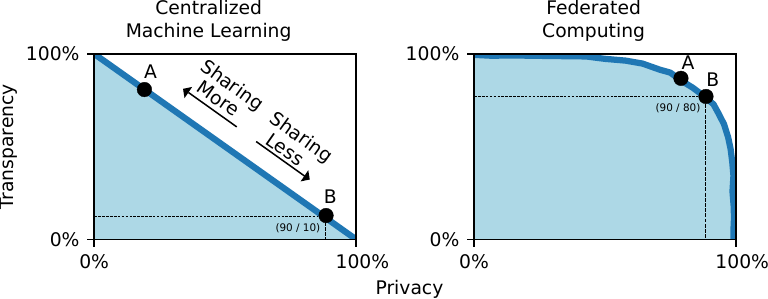}
    \caption{Two Pareto trade-offs between data transparency and privacy. Everything on the curve represents a privacy preserving outcome. The level of privacy for Point B is in both cases the same, but the level of transparency increased in the right trade-off due to privacy-enhancing techniques~\cite{trask2020pareto, priyanshu2021_2_single_manual}.}
    \label{fig:transparency_pareto_tradeoff}
\end{figure}

Privacy enhancing techniques try to reduce data leakages in an information flow. An information flow describes the communication between a server and the clients and what each node is doing with the received information. For example, a client or a server could try to infer information about the underlying raw data with the aggregated computation outputs. All information flows have a trade-off between transparency and privacy~\cite{trask2020pareto, avent2020pareto}. The transparency is high if a server has direct access to the raw data and consequently privacy is low. If no data is shared at all there is a high level of privacy, but no transparency and knowledge gains. Privacy-enhancing techniques allow to keep a certain level of privacy while improving transparency as well. Figure~\ref{fig:transparency_pareto_tradeoff} illustrates this trade-off. The left plot shows a naive trade-off without leveraging any privacy-enhancing techniques. The more a user shares the lower is the privacy level. However, privacy-enhancing techniques, such as the ones mentioned in Table~\ref{tab:privacy_extensions}, allow to keep a certain privacy level and simultaneously increase the transparency or the gained insights.

Data leakages can be clustered into two different categories: Copy problem and bundling problem~\cite{openmined2021copyproblem}. The copy problem describes the loss of control when giving somebody a copy of a dataset. There can be legal boundaries describing to which extend the data can be used. However, enforcing those constraints is challenging. The bundling problem describes information leakages due to an information content, which contains more information than the actual requested one, but they cannot be separated from each other. Therefore, it is possible to directly get additional information, which is not needed or it is possible to do backwards inference to the input data based on the output. An example is the age verification of somebody who wants to buy alcohol. The cashier asks for an ID and verifies that the customer is above the legal age for drinking. However, in the process additional information is leaked such as the name or address of the customer. Those information are not required to complete the age verification process. An example about backward inference is the de-anonymization of some users in a dataset from Netflix movie ratings by comparing rankings and timestamps with public information in the Internet Movie Database~\cite{narayanan2006netflix}. Other examples leveraged anonymized internet usage patterns~\cite{naini2016deanonymize} or location data~\cite{zhang2021deanonymize, luc2019deanonymize} to infer information about individuals. First, we describe FC extensions trying to solve the copy problem and then we focus on solutions to the bundling problem.

\begin{table}[t]
    \caption{\label{tab:privacy_extensions}Privacy enhancing techniques to either protect the input or output against privacy attacks.}
    \begin{tabular}{l | l}
        \textbf{Input Privacy} & \textbf{Output Privacy} \\
        \hline
        Secure-Multi-Party Computation & Differential Privacy \\
        Homomorphic Encryption & Student-Teacher Learning (e.g., PATE~\cite{papernot2018PATE}) \\
        Public-Key Cryptography & \\
        Federated Computing & \\
    \end{tabular}
\end{table}

Input privacy tries to solve the copy problem. Its goal is to keep computation inputs of individuals secret from all parties involved. Table~\ref{tab:privacy_extensions} shows example techniques for achieving input privacy, which mainly originate from the cryptography domain. It is possible to combine them, e.g., FC + homomorphic encryption (HE). This implies that theoretically anybody can run computation tasks on data without the need for direct access to it. This allows an information flow without the need for a trusted third-party.~\cite{trask2020pareto, wang2011privacy}

Secure-Multi-Party Computation (SMPC) enables multiple clients to jointly compute a result without sharing their inputs. They obfuscate their inputs with random numbers, which are randomly distributed to other participating clients. Some disadvantages of this approach are an increase in network traffic due to multiple clients communicating with each other instead of only sending updates to the server and the risk of data loss due to clients dropping out. Most SMPC algorithm rely on a pair-wise collaboration. If one participant of such a pair drops out during the process the added random numbers do not cross out, resulting in a false output.

HE allows to run computations on encrypted data. This approach increases input privacy at the cost of computational complexity. The output of a computation is still encrypted and only the server is able to decrypt the generated results. So there is a trade-off between SMPC's high network overhead or HE's high computational requirements.

Output privacy tries to solve the bundling problem by preventing backwards inference or reverse-engineering of the input based on the output. This is addressed by access control or statistical disclosure control. The former imposes restrictions on who has access to the data. The latter relies on a combination of suppression, perturbation, randomization and aggregation of data~\cite{ricciato2020privacy}. A widely used approach is differential privacy (DP) and related techniques~\cite{trask2020pareto}.

\begin{table}[b]
    \caption{\label{tab:differential_privacy}Core components of a DP algorithm with some examples. An DP algorithm consists of one of each component. The list with examples is not exhaustive.}
    \begin{tabular}{l | c | c}
        \multicolumn{1}{c|}{\textbf{DP Definition}} & \textbf{Randomization Mechanism} & \textbf{Sampling Technique} \\
        \hline
        Pure DP & Gaussian & Poisson \\
        Approximate DP & Laplace & Uniform \\
        Concentrated DP & & \\
        Zero concentrated DP & & \\
        Gaussian DP & & \\
        Rényi DP & & 
    \end{tabular}
\end{table}

DP is a mathematical framework that provides stringent statistical guarantees about the privacy of individuals participating in a database~\cite{dwork2007DP, dwork2008DPSurvey, dwork2011PrivateData}. Its basic idea is to artificially add noise to a data set without changing its statistical properties. It provides provable guarantees about the amount of private data an adversary can infer by observing the outputs of an algorithm. It can either be deployed on client (user-level privacy) or server-side (record-level privacy). Table~\ref{tab:differential_privacy} provides examples for the three core components of each DP algorithm. This summary highlights the complexity of choosing a suitable DP algorithm and it is not an exhaustive list. Several DP definitions exist, with each having different theoretical privacy guarantees. In theory an DP algorithm consists of any combination of those three components.

\subsubsection{Compression} \label{compression}
Running an FC system requires continuous communication between the server and clients. The resulting network traffic can be higher when compared to transferring the raw data. Also, clients' weak network connections might result in dropouts during training. In general, there are two approaches to reduce the network traffic. First, decrease the number of aggregation rounds. Second, reduce the data transfer itself to decrease the overall network traffic. However, both approaches (aggregation frequency and update size) can negatively impact the ML model performance. Therefore, there is a trade-off between those two metrics. Since the upload speed is significantly lower than the download speed, most papers focus only on compressing the gradient updates that clients send each round and leave aside the global server updates. However, in real-world applications, additional servers balance client requests. These so-called parameter servers lead to an increased communication complexity of the global updates. Therefore, an FC system should implement compression in both directions.~\cite{tang2020doublesqueeze}.

\begin{figure}[b]
    \includegraphics{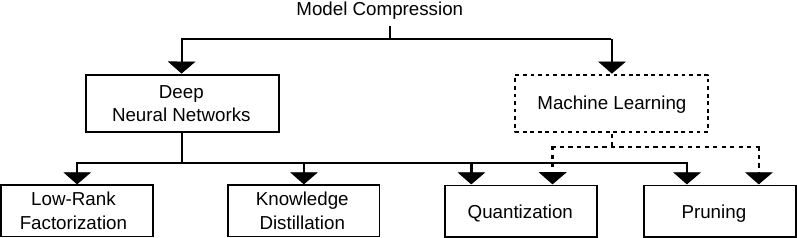}
    \caption{Model compression techniques for deep neural networks and machine learning (e.g., Support-vector-machines and decision trees)~\cite{choudhary2020CompressionSurvey}.}
    \label{fig:compression}
\end{figure}

A server can aggregate after every 10th batch instead of after every batch or decrease the aggregation frequency even further by aggregating after every n-th epoch. However, the longer clients train locally, the more biased they become towards their local dataset. It becomes more challenging for the aggregation server to merge highly personalized models into one general model, which is helpful for all clients. An experimental study with a dataset containing electrical signals used to train four different ML architectures quantifies a reduction in overall model accuracy when increasing the aggregation steps from one batch up to ten batches~\cite{schwermer2022BLONDFl}.

Data compression techniques have a wide range of applications that are not exclusive to FC. However, those extensions boost hardware and network utilization in FC systems. Reducing network requirements makes FC use cases in network constraint environments more feasible. For example, mobile phones running on metered mobile networks or IoT devices connected with narrow bandwidth IoT (NB-IoT) or low range wide area network (LoRaWAN) benefit from smaller updates due to cost and network constraints. NB-IoT and LoRoWAN enable smart city applications due to their long communication range, but they have constraints concerning the maximum package size and energy consumption.

Multiple different model compression techniques exist that reduce storage size (memory or disk), decrease energy consumption for inference, or increase inference speed. Figure~\ref{fig:compression} summarizes four common model compression techniques. Knowledge distillation transforms a large model (teacher) into a smaller and lighter model (student) by letting the student learn from the teacher. The student model tries to learn the generalization capabilities while staying smaller. Quantization leverages different data types with their respective precision. For example, using 32-bit floating-point numbers instead of 64-bit ones saves storage. An advanced version of quantization is the usage of clusters. A cluster consists of either the same values or multiple ranges of values. Instead of storing all numbers, the model only stores those clusters. Some adaptions of quantization in the context of FL are FedPAQ~\cite{reisizadeh2020FLQuantization} and UVeQFed~\cite{shlezinger2020UVeQFed} among others~\cite{shlezinger2020FLQuantization, amiri2020FLQuantization, chen2021FLQuantization}. Model pruning removes weights and neuron connections that are either below a given threshold or do not contribute much to the final result. Model pruning reduces complexity and the number of computations to run. The latter is especially beneficial for edge device scenarios with limited computational resources or battery-powered devices~\cite{jiang2023FLEdgePruning, jiang2022FLEdgePruning, zhang2023FLEdgePruning, yu2023FLEdgePruning}. Lastly, low-rank factorization factorizes a matrix into a product of two matrices with lower dimensions to reduce complexity.~\cite{choudhary2020CompressionSurvey}

\section{Federated Computing Systems} \label{systems_fc}

\begin{table}[b]
    \caption{\label{tab:evaluation_matrix}Example FL systems classified based on our framework. The numbering convention in the motivations columns follows our definitions from Section~\ref{cha:problems} ((1) Improve insights or ML performance, (2) Improve privacy or security, (3) Improve hardware or network utilization)}
    \fontsize{8.9}{11}\selectfont
    \begin{tabular}{c | c | c | c | c | c | c}
        \textbf{Motivation} & \textbf{Environment} & \textbf{Scenario} & \textbf{Framework} & \textbf{Selection} & \textbf{Aggregation} & \textbf{Extension} \\
        \hline
        \multirow{3}{*}{(1)} & \multirow{3}{*}{Single node} & \multicolumn{1}{l|}{model-centric} & \multirow{3}{*}{TFF} & \multirow{3}{*}{Manual} & \multirow{3}{*}{FedAvg} & \multirow{3}{*}{None} \\
            &             & \multicolumn{1}{l|}{horizontal}    &     &        &        &  \\
            &             & \multicolumn{1}{l|}{cross-device}  &     &        &        &  \\
        \hline
        \multirow{3}{*}{(1)} & \multirow{3}{*}{Single node} & \multicolumn{1}{l|}{model-centric} & \multirow{3}{*}{PyTorch} & \multirow{3}{*}{Manual} & \multirow{3}{*}{NaN} & \multirow{3}{*}{None} \\
            &             & \multicolumn{1}{l|}{horizontal}    &     &        &        &  \\
            &             & \multicolumn{1}{l|}{cross-device}  &     &        &        &  \\
        \hline
        \multirow{3}{*}{(3)} & \multirow{3}{*}{Single node} & \multicolumn{1}{l|}{model-centric} & \multirow{3}{*}{NaN} & \multirow{3}{*}{Manual} & \multirow{3}{*}{NaN} & \multirow{3}{*}{None} \\
            &             & \multicolumn{1}{l|}{horizontal}    &     &        &        &  \\
            &             & \multicolumn{1}{l|}{cross-device}  &     &        &        &  \\
    \end{tabular}
\end{table}

Based on our proposed framework in Section~\ref{federated_computing} we categorize a wide range of use case / application papers. Table~\ref{tab:evaluation_matrix} provides a few examples of our literature research. First, we identify the motivation of the paper or the challenges it tries to solve. The numbers (1), (2), and (3) refer to the definitions based on Section~\ref{cha:problems}. The definition of the environment is given in Section~\ref{cha:meta_layer} and we describe the different available scenarios in Section~\ref{scenarios}. The next three columns (framework, client selection, and aggregation strategy) capture the basic building blocks of the FC system. Instead of using an FC specific framework, some papers leverage ML framework such as \textit{PyTorch} or \textit{TensorFlow}, or they do not specify the used framework at all. A NaN indicates missing information about the framework or aggregation strategy. The first row in the table uses the FL framework \textit{TFF}, which stands for \textit{TensorFlow Federated}. All three examples in Table~\ref{tab:evaluation_matrix} run without any kind of privacy-enhancing or compression extension. Lastly, we summarize the implemented extensions. With our framework we capture an FC system completely, with all its unique components. We visualize our findings and describe them in detail in the following sections. The goal is to identify FL and FA system configurations, which are either extensively implemented in the research community or are lacking further investigations. This enables the identification of research trends and gaps.

\subsection{Federated Learning Systems} \label{systems_fl}

\begin{table}[b]
    \centering
    \caption{Categorization of FL papers into three groups. A reference can appear in multiple groups. Environment describes the number of hardware used for an experiment. Motivation refers to three categories introduced in Section~\ref{cha:problems} ((1)) = Improve insights or ML performance, (2) = Improve privacy or security, (3) = Improve hardware or network utilization). Client selection is either done manually, resource-aware, or loss-aware.}
    \begin{tabular}{l | p{0.203\columnwidth} | p{0.203\columnwidth} | p{0.203\columnwidth} | p{0.203\columnwidth}}
        \multicolumn{2}{l|}{\diagbox[width=11em]{\textbf{Motivation}}{\textbf{Environment}}} & \multicolumn{1}{c|}{\textbf{Unknown}} & \multicolumn{1}{c|}{\textbf{Single Node}} & \multicolumn{1}{c}{\textbf{Multiple Nodes}} \\
        \hline
         \multirow{8}{*}{(1)} & Manual \newline Selection &\cite{lee2022_1_unknown_manual, tun2021_1_unknown_manual, liu2020_1_unknown_manual, li2022_123_unknown_manual, thorgeirsson2021_1_unknown_manual, yang2021_1_unknown_manual, feki2021_1_unknown_manual, lai2022_1_unknown_manual, abidin2022_1_unknown_random, mushtaq2023_1_unknown_manual, chuanxin2020_12_unknown_manual} & \cite{taik2020_1_single_manual, guo2020_1_single_manual, chen2022_1_single_manual, du2020_1_single_manual, saputra2019_1_single_manual, ye2020_1_single_manual, zhu2019_1_single_manual, li2021_1_single_manual, wang2021_1_single_manual, wang2021_1_single_manual, dai2023_1_single_manual, ye2020_1_single_manual, sarma2021_1_single_manual, abdul2021_1_single_manual, qiu2023_1_single_manual, zou2022_1_single_manual, zou2022_1_single_manual_, farcas2022_13_manual_single, kortoci2022_13_manual_single} & \cite{zhang2023FLEdgePruning, rajendran2021_1_multiple_manual, schwermer2022_13_multiple_manual, aloqaily2022_1_multiple_manual, houda2022_1_multiple_manual, korkmaz2022_13_multiple_manual, pan2022_1_multiple_manual, arouj2022_13_multiple_manual} \\ \cline{2-2}
         & Resource-aware \newline Selection & \multirow{2}{*}{\cite{xu2021_13_unknown_resource}} & \multirow{2}{*}{\cite{nishio2019_13_single_resource}} & \\ \cline{2-2}
         & Loss-aware \newline Selection & \multirow{2}{*}{\cite{cho2020_13_unknown_loss}} & & \\
        \hline
         \multirow{2}{*}{(2)} & Manual & \multirow{2}{*}{\cite{li2022_123_unknown_manual, chuanxin2020_12_unknown_manual}} & \multirow{2}{*}{\cite{priyanshu2021_2_single_manual}} & \\
         & Selection & & & \\ \cline{2-2}
        \hline
         \multirow{6}{*}{(3)} & Manual \newline Selection & \multirow{2}{*}{\cite{li2022_123_unknown_manual}} & \cite{yang2020_3_single_manual, wang2020_3_single_manual, zeng2020_3_single_manual, zhao2022_3_single_manual, farcas2022_13_manual_single, kortoci2022_13_manual_single} & \cite{qiu2020_3_multiple_manual, mo2021_3_multiple_manual, schwermer2022_13_multiple_manual, schwermer2023_3_multiple_manual, korkmaz2022_13_multiple_manual, arouj2022_13_multiple_manual} \\ \cline{2-2}
         & Resource-aware & \multirow{2}{*}{\cite{xu2021_13_unknown_resource}} & \multirow{2}{*}{\cite{nishio2019_13_single_resource}} & \multirow{2}{*}{\cite{abdulrahman2021_3_multiple_resource}} \\
         & Selection & & & \\ \cline{2-2}
         & Loss-aware & \multirow{2}{*}{\cite{cho2020_13_unknown_loss}} & & \\
         & Selection & & & \\
    \end{tabular}
    \label{tab:fl_survey}
\end{table}

The focus in research is currently on FL. It covers applications from all domains, such as energy, mobility, and healthcare. Table~\ref{tab:fl_survey} provides an excerpt of our literature research for FL systems. It shows all surveyed papers, but not all captured characteristics. Papers can appear multiple times if they have more than one motivation. For example, a paper can improve the ML performance and simultaneously try to improve hardware utilization. The y-axis in Table~\ref{tab:fl_survey} contains the motivation and client selection approaches. Section~\ref{cha:problems} describes each of those motivations in detail. We separate client selection strategies into manual, resource-aware, and loss-aware. In the first strategy, the authors either use all available clients or manually define a client set for training. The other two client selection strategies define a client set for training based on the available resources on the client side or on how the loss behaves during a training round. The x-axis shows the experiment environment. It is unknown when the publication does not state the hardware used for the experiments. If the publication contains information about the hardware, it is either a single node hosting the server and clients or multiple nodes. The latter allocates one piece of hardware for the server and each client, respectively. The hardware includes edge devices, such as Raspberry Pi, dedicated servers, or GPUs. The distribution of a multi-node environment is either located in one location or spatially distributed.

\begin{figure}[b]
    \includegraphics{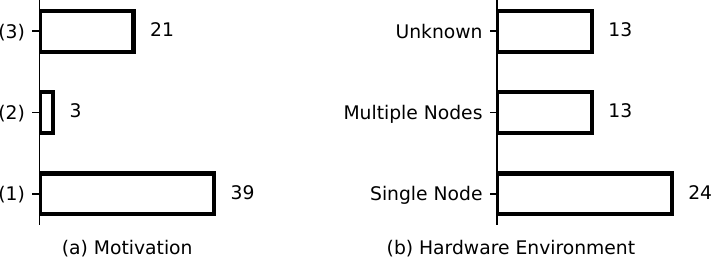}
    \caption{Distributions of the problem definition (a) and hardware environment (b) of summarized FL systems. The numbers follow the structure introduced in Section~\ref{cha:problems} ((1) Improve insights or ML performance, (2) Improve privacy or security, (3) Improve hardware or network utilization).}
    \label{fig:barplots_FL}
\end{figure}

The majority of publications focus on improving machine learning prediction performance by running experiments on either an unknown system or an environment with all parties (server and clients) being simulated on one device. A one-device approach with simulated clients reduces the complexity and overhead of the system. Having a realistic network traffic or monitoring the CPU cycles are not necessary to investigate the effect of different ML and aggregation strategies on the final models' accuracy. A major challenge of FL systems is an uneven distribution of labels on the clients (non-IID). Therefore, it makes sense to first focus on developing ML prototypes which achieve satisfactory accuracy before increasing the systems' complexity by introducing hardware, network, or energy constraints. Most papers do not specify which environment they run on (unknown). Figure~\ref{fig:barplots_FL} shows the distribution of motivation (a) and experiment environment (b) for the surveyed FL systems. 49 out of 50 papers focus on improving machine learning prediction performance. Eight of those 49 papers also try to improve hardware or network constraints. The next prevalent motivation is to improve hardware or network utilization (motivation (2) in Figure~\ref{fig:barplots_FL}). To capture changes in those metrics, it is paramount to run the experiments on distributed devices. However, only 26~\% of FL papers run their experiments on multiples nodes. Papers using one node (48~\%) or an unknown environment (26~\%) often test an optimization function with respect to hardware or network utilization improvements. So, instead of measuring actual utilization rates, they theoretically estimate them.

The preferred client selection algorithm and aggregation strategy are manual and \textit{FedAvg}, respectively. There is also a clear trend towards using unspecified (unknown) aggregation strategies 12~\%) or \textit{FedAvg} (59~\%). The latter is a simple aggregation strategy, and it is mainly used in its pure form. There is a huge variation of aggregation strategies. Our survey captures at least 11 different strategies, and the majority of publications work with only one aggregation strategy. Seven publications deployed at least one aggregation strategy. We visualize its distribution in Figure~\ref{fig:barplots_FL} by counting all occurrences independent of its publication. Only a few publications adopt aggregation strategies developed in other work. For example, only two publications leverage \textit{FedAvgM}. Reasons for a lack of wide adaptations of new aggregation strategies, besides \textit{FedAvg}, are manifold. The developed aggregation strategy might be to specialized for a specific use case or dataset. Other reasons can be a lack of documentation or source code. Having a standardized way of describing an aggregation strategy highlights the differences and advantages of specific strategies. The summary in Table~\ref{tab:aggregation_strategies} in Section~\ref{cha:aggregation_algorithm} is a starting point for expanding it to more detailed definitions of the in- and outputs of each aggregation strategy to better reproduce and understand its functionality. 

Also, 94~\% of our surveyed FL papers capture the same scenario, which represents a model-centric, horizontal FL and cross-device architecture. The exceptions use cross-silos instead of cross-devices. They come from the healthcare domain and capture different institutions instead of multiple devices. A cross-silo architecture is similar to a cross-device one. Differences between those two architectures are rather on legal basis instead of a technological one. However, implementing data-centric instead of model-centric or vertical FL instead of horizontal FL introduces new challenges and increases the overall complexity. Data-centric architectures improve the ML model accuracy from an FL system by altering a clients' dataset and vertical FL merges datasets with different features. Therefore, currently the focus is on model-centric and horizontal FL systems to keep the complexity low. However, this highlights a lack of research in more complex FL systems incorporating datasets with different feature sets.

\begin{figure}[t]
    \includegraphics{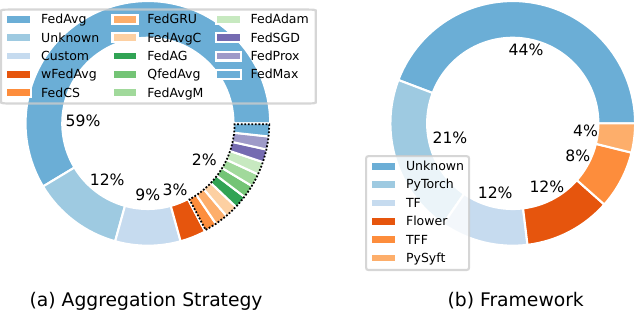}
    \caption{Distributions of aggregation strategies (a) and frameworks (b) for FL papers. We surveyed in total 50 FL papers published after 2020.}
    \label{fig:pies_fl}
\end{figure}

Figure~\ref{fig:pies_fl} provides an overview of deployed frameworks and aggregation strategies. Most FL papers do not specify the framework used (44~\%), or they leverage frameworks widely used in ML applications, such as  \textit{PyTorch} (21~\%) or \textit{TensorFlow} (12~\%). FL-ready frameworks with an integrated communication and aggregation layer are the minority. Only 24~\% of the papers use an FL framework, such as \textit{TensorFlow Federated} (4~\%), \textit{PySyft} (8~\%), or \textit{Flower} (12~\%). We introduce a wide range of FL frameworks in Section~\ref{frameworks_and_network_protocols}. However, almost none achieved a wide range adaptation due to usage/license constraints or too short update cycles. For example, the FL framework from IBM has a community and enterprise edition, and only the latter is open for commercial use. Such constraints hinder adaptation. Additionally, their last version is almost 1.5~years old. Another reason for not using a specific FL framework is the ease of use. \textit{Flower} only requires the installation of its \textit{Python} package whereas \textit{PySyft} requires Docker and a local database. All FL frameworks have some tutorials or installation guides, but more requirements increase complexity and potential sources of errors. Therefore, there seems to be a trend towards lean FL frameworks such as \textit{TensorFlow Federated} and \textit{Flower}, which use optimized serialization (Protobuf) and communication protocols (gRPC). Nevertheless 76~\% of publications are most likely not built for a real-world FL deployment because they rely purely on ML frameworks, such as \textit{TensorFlow} or \textit{PyTorch}. The majority of publications using such ML frameworks focus on improving ML training in an FL system. Therefore, we infer that all publications with an unknown framework focusing also on ML performance use ML frameworks as well.

\subsection{Federated Analytic Systems} \label{systems_fa}
The current driving motivation for FA is similar to FL systems' main motivation: Improving generated insights. FA systems run for one round and do not have an iterative optimization approach. Therefore, the focus is not on improving ML prediction performance, but rather on generating and aggregating single performance metrics. The captured scenarios for FL and FA are also similar and both focus on model-centric, horizontal FC in a cross-device environment. No surveyed FA paper works with either data-centric, vertical or cross-silo environments and the publication year for all publications are not older than 2021. This highlights the recent research interest in FA. Table~\ref{tab:fa_survey} provides an overview of our literature research.

\begin{table}[t]
    \centering
    \caption{Categorization of FA papers into three groups. A reference can appear in multiple groups. Environment describes the number of hardware used for an experiment. Motivation refers to three categories introduced in Section~\ref{cha:problems} ((1)) = Improve insights or ML performance, (2) = Improve privacy or security, (3) = Improve hardware or network utilization). Client selection is either done manually, resource-aware, or loss-aware.}
    \begin{tabular}{l | p{0.203\columnwidth} | p{0.203\columnwidth} | p{0.203\columnwidth} | p{0.203\columnwidth}}
        \multicolumn{2}{l|}{\diagbox[width=11em]{\textbf{Motivation}}{\textbf{Environment}}} & \multicolumn{1}{c|}{\textbf{Unknown}} & \multicolumn{1}{c|}{\textbf{Single Node}} & \multicolumn{1}{c}{\textbf{Multiple Nodes}} \\
        \hline
         \multirow{2}{*}{(1)} & Manual \newline Selection &\cite{hamersma2023FA_1_unknown_manual, froelicher2021FA_1_unknown_manual, chen2021FA_1_unknown_manual, gjoreski2022_12_unknown_manual, wang2023FA_1_unknown_manual, wang2023FA_13_unknown_custom, cormode2021FA_12_unknown_unknown, koga2023FA_12_unknown_unknown} & \multirow{2}{*}{\cite{wang2022FA_1_single_random, pandey2022_13_single_manual, bagdasaryan2022_123_single_manual}} & \multirow{2}{*}{\cite{zhang2022FA_13_multiple_random, margolin2023FA_123_multiple_unknown}} \\ \cline{2-2}
        \hline
         \multirow{2}{*}{(2)} & Manual & \multirow{2}{*}{\cite{gjoreski2022_12_unknown_manual, chaulwar2021FA_2_unknown_unknown}} & \multirow{2}{*}{\cite{bagdasaryan2022_123_single_manual, elkordy2022FA_23_single_unknown, cormode2021FA_12_unknown_unknown, koga2023FA_12_unknown_unknown}} & \multirow{2}{*}{\cite{short2022_2_multiple_manual, shi2022_2_multiple_manual, margolin2023FA_123_multiple_unknown}} \\
         & Selection & & & \\ \cline{2-2}
        \hline
         \multirow{2}{*}{(3)} & Manual & \multirow{2}{*}{\cite{wang2023FA_13_unknown_custom}} & \multirow{2}{*}{\cite{pandey2022_13_single_manual, bagdasaryan2022_123_single_manual, zhao2023FA_3_single_manual, toka2023FA_3_single_manual, elkordy2022FA_23_single_unknown}} & \multirow{2}{*}{\cite{zhang2022FA_13_multiple_random, margolin2023FA_123_multiple_unknown}} \\ 
         & Selection & & & \\
    \end{tabular}
    \label{tab:fa_survey}
\end{table}

The type of hardware environment used for FA use cases is more evenly distributed when compared to the FL systems. Running the experiments on an unknown environment, a single node, or multiple nodes associate for 21~\%, 32~\%, and 47~\%, respectively. There is no clear match between motivation and environment. For example, papers looking at hardware or network metrics use 63~\% of the times a single node and only 25~\% of those papers deploy their system on multiple nodes. This looks a bit different for papers capturing privacy or security motivations. About 45~\% of them use an unknown environment, whereas 22~\% and 33~\% run their experiments on a single node or multiple nodes, respectively. The client selection is either manual, random, or unknown. 53~\% of the papers select clients manually and 16~\% use a random client selection. For both approaches the set of clients stays constant during the training. A manual client selection describes either an undefined client selection or all available clients are always selected. The latter could be a dataset which is inherently non-IID. For example, electricity measurements from multiple households which have a different dominant label per house (e.g., TV for household 1 and washing machine for household 2). However, 31~\% of papers do not specify the type of client selection.

\begin{figure}[t]
    \includegraphics{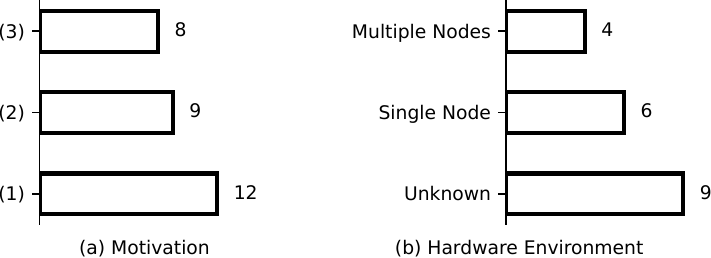}
    \caption{Distributions of the problem definition (a) and hardware environment (b) of summarized FA systems. The numbers follow the structure introduced in Section~\ref{cha:problems} ((1) Improve insights, (2) Improve privacy or security, (3) Improve hardware or network utilization).}
    \label{fig:barplots_FA}
\end{figure}

Figure~\ref{fig:barplots_FA} shows the distribution of frameworks and aggregation strategies used for FA systems. Not a single paper uses a framework specifically built for FA or FL and 74~\% of papers do not mention which framework they use at all. The majority of papers use an ML framework, such as \textit{TensorFlow}, \textit{PyTorch} or \textit{MATLAB}, to simulate an FA environment. It is feasiable to use an adjusted version of a framework dedicated for FL systems in an FA system. This requires only little to none changes, because in the simplest scenario an FL framework runs for only one round to mimic an FA system. This enables researchers to leverage the existing communication and serialization infrastructure. Therefore, it is not clear why no FA paper uses existing FL frameworks even when they try to improve hardware and network utilization. The field of frameworks used in FA systems is also less divers when compared to FL systems (three vs. five frameworks). However, the ratio of unknown frameworks for FA systems is 30~\% percentage points higher, which introduces uncertainties when comparing FA with FL systems.

\begin{figure}[b]
    \includegraphics{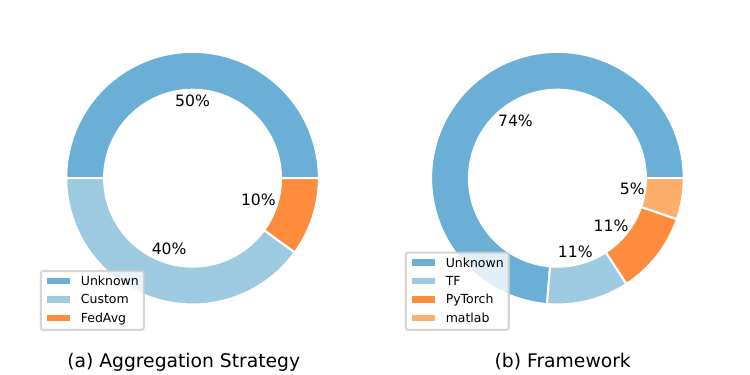}
    \caption{Distributions of aggregation strategies (a) and frameworks (b) for FA papers. We surveyed in total 19 FA papers published after 2020.}
    \label{fig:pies_fa}
\end{figure}

Figure~\ref{fig:pies_fa} highlights the distribution of aggregation algorithms. Aggregation strategies are also mostly unknown (50~\%) and the focus is currently on custom strategies (40~\%). They often work with privacy-enhancing algorithms, such as homomorphic encryption, differential privacy or secure-multi party aggregation. The number of different aggregation strategies is much lower when compared to FL systems. A reason could be the reduced complexity of FA systems and the smaller impact of aggregation strategies on the final result, because an FA training runs for one round instead of multiple ones with an optimizer running on each client, which adds an additional layer of complexity.

However, half of the surveyed papers extend FA with privacy enhancing techniques. The privacy aspect is more present for FA systems compared to FL ones. FA training runs for only one round and often with deterministic models. The risk of de-annonymizing or reverse-engineering the raw data based on the clients' output is higher compared to statistical models with some degree of randomness. Therefore, it makes sense that about 35~\% of surveyed papers combine FA systems with privacy-enhancing strategies. Improving generated insights is also easier compared to FL systems due to the reduction of system complexity. This allows researchers to focus on other topics.

\section{Discussion} \label{discussion}

\subsection{Current Research Trends} \label{current_research_trends}
FL got attention in the industry and academia starting in 2016 and FA emerged in 2020. However, their precursor such as Federated Databases have been studied long before that. Its main applications are in the ML domain. Research focuses currently on improving ML prediction performance by introducing new client selection (e.g., resource-aware or loss-aware client selection), aggregation algorithms or fine tuning hyperparameters of ML models. The primary challenge is a non-IID of labels on client-side leading to biased and unbalanced model updates. The network traffic and hardware consumption are for those experiments neglectable. Therefore, to test new ML-focused approaches, it is sufficient to deploy experiments on one machine which contains the server and multiple simulated clients. This strategy also reduces complexity and organizational overhead when compared to experiments running on distributed systems. The second most found motivation is improving hardware and network utilization of FL systems. Those papers solely focus on this issue. However, a few combine improving ML and hardware performance. Papers also considering the hardware or network utilization of an FL system run about half and half on either a single machine or on multiple ones. A few experiments investigate the impact of FL on those metrics theoretically by compiling optimization functions with constraints. Lastly, only a few papers extend FL systems with additional privacy-enhancing techniques, such as multi-party computation or DP. On the other side, papers describing FA systems mainly combine it with some kind of privacy-enhancing techniques. An explanation is the inherently deterministic nature of FA due to its lack of any optimizations running on client side. Therefore, it is easier to reverse-engineer raw data based on each clients individual update. To counter this, more works combine FA with techniques improving either input or output privacy.

Another focus in research is on model-centric, horizontal, and cross-device FC. This is the simplest scenario for FL and FA. Its goal is to improve ML performance by tuning hyper parameters, which is easier with a pre-defined homogeneous set of features and devices. The distribution of features per client can vary widely, but the feature set is the same for all of them. On the other side, data-centric and vertical FL and FA increases complexity. A data-centric approach aims at increasing ML performance by improving data quality. However, working with multiple distributed clients with a divers level of ownership makes it challenging to make changes on client-side. Also, vertical FL and FA work with multiple different feature sets, which can also vary per client.

About 99~\% of surveyed papers incorporate a centralized architecture (see Figure~\ref{fig:fc_architectures}). Only one looks at the advantages and challenges of hierarchical~\cite{bonawitz2019FLSystemDesign} and none at peer-to-peer systems.

\subsection{Open Challenges} \label{open_challenges}
FL and FA often lack information about deployed systems, such as environment, framework, or aggregation strategy, which decreases reproducibility. Our proposed standardized framework to describe FC systems enables other researchers and stakeholders to identify similar systems for comparison. 

FL and FA systems run mostly on one node or the environment is unknown due to the focus on ML accuracy improvements. This leads to a lack of understanding on how FC systems perform under real-world scenarios and how certain client selection and aggregation strategies affect network traffic and other hardware metrics, such as CPU utilization or energy consumption. Running an experiment on one device neglects potential latency or throughput bottlenecks of the network or the hardware itself. Therefore, ML-focused papers should be as reproducible as possible to enable other researchers to quantify their impact on the above mentioned metrics and to optimize hardware utilisation during the training process. Orchestration of multiple nodes in a real-world FC scenario is also not well researched.

Client selection algorithms primarily consider all available clients and the client set stays constant during training. Most dataset used in FC papers have a limited number of clients and reducing them could lead to an insufficient amount of (training) data. However, resource-aware or loss-aware client selection algorithms could improve ML performance and hardware utilization. Also, re-selecting a new subset of clients after an ML round increases complexity, but could potentially improve the entire systems' performance.

Not much work for FA exist. Strategies working for FL systems might achieve similar results in an FA system. FL-specific strategies, such as loss-aware client selection is not applicable in an FA context, because FA runs for one run and hence, cannot incorporate iterative optimizations. There is also a lack of using FC-specific frameworks in an FA system.

\section{Conclusions} \label{conclusion}
FL and FA enable the development of data-driven business models by leveraging data silos without interfering with data protection laws and by eliminating reservations from decision makers and stakeholders. Such models are paramount for improving business processes and to cope with the ever increasing complexity and velocity of changes in the global economy. Both approaches belong to FC. Over the last years researchers and practitioners expand FC systems with algorithms from other domains (e.g., encryption and compression) to minimize the effect of some of its disadvantages. However, those systems become more complex and there is currently a lack of clearly defined boundaries for such systems. Our work introduces a taxonomy capturing all moving parts in an FC systems. We differentiate between FL and FA. We summarize current research trends in FC systems and identify gaps. Additionally, we categorize existing frameworks and client selection algorithms. Currently, the majority of publications focus on FL systems with the goal to improve the accuracy of the trained machine learning models. The experiments mainly run on one device hosting the server and clients. There is  a lack of research on the effect of FL Training on hardware utilization. A similar picture exists for FA systems. However, publications in this area tend to incorporate more privacy-enhancing techniques, such as differential privacy or secure-multi party computation. Our taxonomy servers as a blue print for further research on FC systems. Additionally, our comprehensive summary of existing FL frameworks highlights the focus on the combination of gRPC with Protobuf for communication and object serialization. 

\begin{acks}
This paper is funded by the Bavarian Ministry of Economic Affairs, Regional Development and Energy under the REMORA project with the identifier DIK0111/03.
\end{acks}

\bibliographystyle{ACM-Reference-Format}
\bibliography{references}

\appendix

\end{document}